\documentclass[journal]{IEEEtran}

\usepackage{booktabs}
\usepackage{multirow}
\usepackage{colortbl}
\usepackage{xcolor}
\usepackage{hyperref}

\usepackage{amsmath,amssymb,amsfonts}
\usepackage{graphicx}
\usepackage[table]{xcolor} 
\usepackage{url}
\usepackage{cite}
\usepackage{tikz}
\usepackage{algcompatible}
\usepackage{algorithm}
\usepackage{multirow}
\usepackage{pifont}
\usepackage{xspace}

\usepackage{enumitem}
\usepackage[utf8]{inputenc}
\usepackage{xspace}

\usepackage{microtype}
\usepackage{booktabs}
\usepackage[justification=centering]{caption}
\usepackage{makecell}
\usepackage{tabularx}
\usepackage{array}
\usepackage{orcidlink}
\usepackage{float}          
\usepackage{caption}        

\definecolor{highlightblue}{RGB}{235,235,255}
\definecolor{colOrig}{RGB}{255,240,240}
\definecolor{col1080p}{RGB}{240,255,240}
\definecolor{col4K}{RGB}{240,240,255}



\title{RoofGS: Roofline-Guided End-to-End Acceleration of 3D Gaussian Splatting}

\author{
    Yang Luo\orcidlink{0009-0002-7124-2296}, 
    Yan Gong\orcidlink{0000-0002-3148-8286}, 
    Yongsheng Gao*\orcidlink{0000-0002-1555-8328}, 
    Jie Zhao\orcidlink{0000-0002-6086-9387},~\IEEEmembership{Senior Member,~IEEE}

\thanks{This work was supported by the National Science and Technology Major Project (Grant No. 2025ZD1603200) and the National Outstanding Youth Science Fund of the National Natural Science Foundation of China (Grant No. 52025054). (\textit{Corresponding author: Yongsheng Gao})}%
\thanks{Yang Luo, Yan Gong, Yongsheng Gao, and Jie Zhao are with the State Key Laboratory of Robotics and Systems, Harbin Institute of Technology, Harbin 150001, China (email: christoluo@outlook.com; gongyan2020@foxmail.com; gaoys@hit.edu.cn; jzhao@hit.edu.cn).}%
}

\begin{document}

\maketitle

\begin{abstract}
3D Gaussian Splatting (3DGS) enables real-time novel-view synthesis but remains limited on GPUs at high resolutions. Through a stage-wise Roofline characterization, we identify two distinct hardware bottlenecks: global memory traffic dominates the front end, whereas instruction throughput limits rasterization. Guided by this analysis, we develop RoofGS, a rendering framework that applies bottleneck-specific optimizations rather than generic kernel acceleration. For the memory-bound front end, we design a resolution-adaptive quantized depth sorting key that compresses each key to 32 bits. For the compute-bound rasterizer, we introduce a range-aware bit-level fast exponential approximation tailored to the bounded exponent range after opacity culling, with a derived per-pixel error bound. These two core techniques are complemented by additional optimizations (kernel fusion, compact attribute storage, culling, dual-pixel evaluation) that additionally reduce memory traffic and improve instruction-level parallelism. Experiments show that RoofGS achieves a 10.1$\times$ end-to-end speedup over 3DGS at 4K on an RTX 4090, increasing throughput from 61 to 616 FPS, with only a 0.028 dB PSNR loss.
\end{abstract}

\begin{IEEEkeywords}
3D Gaussian Splatting, Hardware optimization, Low-bit quantization, Rasterization, Roofline model.
\end{IEEEkeywords}

\section{Introduction}
\label{sec:intro}
Novel view synthesis has emerged as an important technique for applications such as augmented and virtual reality (AR/VR), immersive gaming, robotic perception, and digital twins\cite{survey2024, 3dgsSurvey2025, robot_perception_2025, RenderingReview2026, VINGS-Mono_2025}. Among recent approaches, 3D Gaussian Splatting (3DGS)\cite{Kerbl20233DGS} represents a scene using a set of anisotropic Gaussian primitives and renders novel views through a rasterization-based pipeline. Compared with NeRF-based methods~\cite{mildenhall2020nerf}, 3DGS generally provides faster optimization and rendering while achieving comparable or higher visual quality. However, its rendering cost depends on factors such as the number of Gaussian primitives, image resolution, and scene complexity. For large-scale scenes or high-resolution rendering, the associated computation, memory usage, and data movement remain substantial, hindering the practical deployment of 3DGS in real-time applications.

\begin{figure}
    \centering
    \includegraphics[trim={4.5cm 4cm 4cm 2cm}, clip, width=\linewidth]{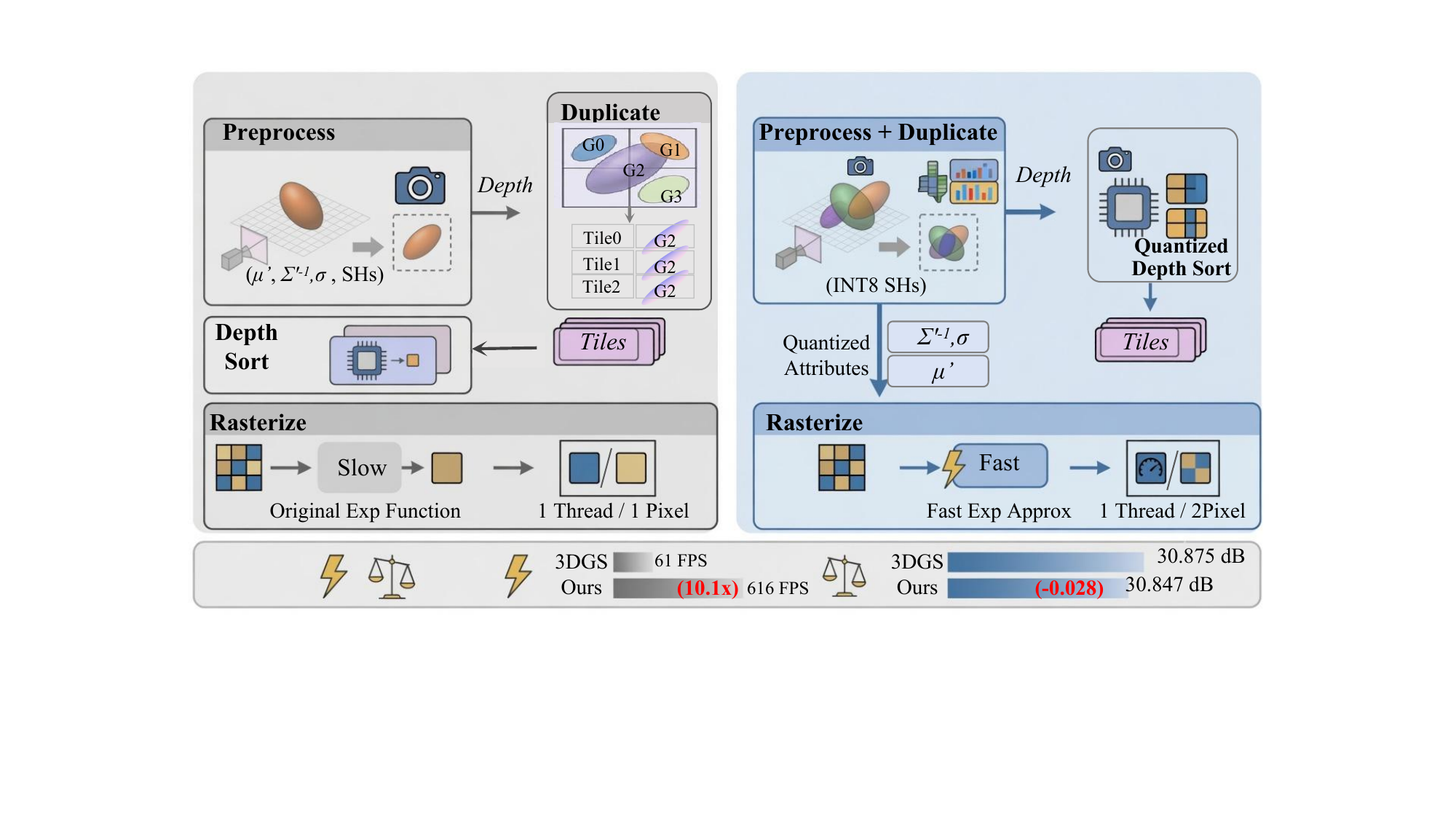}
    \captionsetup{
        font=small,
        labelfont=bf,
        justification=justified,
    }
    \caption{Rendering pipeline comparison. RoofGS fuses preprocessing and duplication, employs INT8 spherical harmonics and quantized depth sorting, and
accelerates rendering through fast exponential approximation and dual-pixel
parallelism.}
    \label{fig:rendering_comparison}
\end{figure}


Existing approaches to accelerating 3DGS rendering mostly follow a
stage-specific optimization paradigm. TC-GS~\cite{liao2025tcgs} and
GEMMGS~\cite{li2026GEMMGS} accelerate rasterization using Tensor Cores,
whereas SpeedySplat~\cite{speedy-splat} and AdR-GS~\cite{wang2024adr} primarily focus on reducing front-end overhead.
Although effective, these methods optimize individual stages or operations,
and their local gains may not translate into proportional end-to-end
acceleration when the bottleneck shifts across the pipeline.

Some studies, particularly hardware-oriented analyses, have also observed
that the stages of the 3DGS rendering pipeline exhibit substantially different
execution characteristics~\cite{lee2024gscore, pei2025gcc, STREAMINGGS_2025}.
The front end, including Gaussian projection, frustum culling, tile assignment
and duplication, and depth sorting, typically exhibits irregular memory-access
patterns and low arithmetic intensity. Rasterization, by contrast, involves
more compute-intensive per-pixel operations, including Gaussian evaluation,
exponential computation, and alpha blending~\cite{Duplex-GS_2026,
lee2024gscore}. However, these observations have been explored mainly in hardware-oriented
contexts, with limited attention to coordinated software
optimization across the whole rendering pipeline.

We present RoofGS, a Roofline-guided implementation framework for end-to-end 3DGS rendering acceleration. As illustrated in Fig.~\ref{fig:rendering_comparison}, RoofGS uses stage-wise performance characterization to match optimization strategies to the dominant hardware constraints of the baseline pipeline, by reducing intermediate data movement in the front end and reducing arithmetic overhead in rasterization. In the front end, it compresses SH and geometric attributes, packs the tile index and quantized depth into a dynamic 32-bit sorting key, and fuses preprocessing, tile duplication, and key generation into a single kernel. During rasterization, RoofGS evaluates two adjacent pixels and replaces the standard exponential operation with a bit-level approximation tailored to alpha-thresholded blending, thereby reducing arithmetic and special-function-unit (SFU) pressure.
 At 4K resolution on an NVIDIA RTX 4090 GPU, RoofGS increases the average rendering throughput from 61 FPS to 616 FPS across Mip-NeRF 360~\cite{Mip-NeRF-360}, Tanks \& Temples~\cite{knapitsch2017tanks}, and Deep Blending~\cite{hedman2018deep}, delivering about $10\times$ end-to-end speedup with an average PSNR reduction of only 0.028 dB.

 The main contributions are summarized as follows:

\begin{itemize}
    \item We present a Roofline-guided stage-wise optimization methodology that allocates optimization effort according to the bottleneck of each stage, rather than applying uniform kernel acceleration.
    
    \item We design a resolution-adaptive quantized depth sorting key that compresses the conventional 64-bit key into 32 bits, reducing sorting traffic and the number of radix passes.
    
    \item We introduce a range-aware bit-level fast exponential approximation tailored to the bounded exponent range after opacity culling, together with a derived per-pixel error bound.
    
    \item We implement and evaluate RoofGS on multiple datasets and GPUs, achieving $10\times$ end-to-end acceleration with negligible quality loss.
\end{itemize}

\section{Related Work}
\label{sec:related}
\subsection{3D Gaussian Splatting}

3D Gaussian Splatting (3DGS)~\cite{Kerbl20233DGS, 3dgsSurvey2025, Baosurvey2025} represents 3D scenes using millions of optimizable Gaussian primitives, each parameterized by its spatial position, covariance, opacity, and view-dependent color encoded via Spherical Harmonics (SHs) coefficients. As illustrated on the left of Fig.~\ref{fig:rendering_comparison}, the standard pipeline routes these attributes through distinct preprocessing and rasterization stages. In the \textit{Preprocess} stage, the 3D Gaussians are projected onto the camera image plane to compute their 2D means $\boldsymbol{\mu}'$ and inverse 2D covariance matrices $\Sigma'^{-1}$. In the \textit{Duplicate} stage, primitives that intersect multiple tile boundaries are replicated across all overlapping tiles, while recording their camera-space depth $D$. The \textit{Depth Sort} stage then groups these duplicated primitives by tile ID and sorts them within each tile by depth in back-to-front order. Finally, the \textit{Render} stage feeds the sorted lists into a tile-based rasterizer. Each tile is typically processed by a thread block that performs alpha blending over the contributing Gaussians to accumulate pixel colors. This design enables efficient parallel rendering but can still incur substantial memory bandwidth and computational overhead at high resolutions such as 4K due to frequent attribute streaming and per-Gaussian evaluations.

\subsection{Acceleration of 3DGS}
Existing acceleration strategies improve 3DGS performance from three complementary perspectives: kernel-level optimization, pipeline-level optimization, and representation-level optimization.

\textbf{Kernel-level optimization} accelerates individual computational hotspots without fundamentally changing the rendering pipeline. Representative culling methods, including Speedy-Splat~\cite{speedy-splat}, AdR-GS~\cite{wang2024adr}, Fast-GS~\cite{ren2025fastgs}, OccluGaussian~\cite{OccluGaussian_2025} and Proxy-GS~\cite{Gao_2026_CVPR}, reduce redundant primitives using culling strategy. Rendering-oriented approaches such as GEMMe-GS~\cite{li2026GEMMGS} and TC-GS~\cite{liao2025tcgs} reformulate Gaussian evaluation to better exploit GPU Tensor Cores. While effective for their target kernels, these methods leave the remaining pipeline unchanged. Consequently, memory-bound stages, including preprocessing, attribute streaming, and sorting, continue to dominate execution time, limiting end-to-end acceleration.

\textbf{Pipeline-level optimization} jointly redesigns multiple rendering stages to improve overall throughput. Existing systems such as FlashGS~\cite{FlashGS2025}, StopThePop~\cite{radl2024stopthepop}, SEELE~\cite{zhu2026seele}, and Mobile-GS~\cite{du2026mobilegs} optimize dataflow, memory access, and execution scheduling to reduce synchronization overhead and improve data locality. However, these designs are largely empirical and do not explicitly link their optimizations to the distinct hardware limits of each stage, making it difficult to decide how to balance memory-traffic reduction and arithmetic-cost reduction across scenes.

\textbf{Representation-level optimization} reduces representation size~\cite{compressionsurvey} through pruning~\cite{wang2026prune, fan2024lightgaussian, Compressed3DGS2024, PUP_3D-GS2025, POTR_2026, Visibility2026, Adversarial_2026}, quantization~\cite{Compact3d2024, HAC++2025, Xuquantization2025}, and distillation~\cite{distill3dgs, Labe2024DGD, xie2024mesongs}, at the cost of additional retraining or post-training. These methods reduce storage requirements and offline memory consumption, but aggressive compression may degrade fine geometry and high-frequency textures. Moreover, existing quantization methods~\cite{Compact3d2024, Xuquantization2025, HAC++2025} primarily focus on storage compression. During rendering, the quantized Gaussian attributes are typically dequantized to floating-point values, which limits the potential runtime benefits for operations such as attribute streaming, projection, and sorting.

Although these paradigms address complementary aspects of the rendering pipeline, limited attention has been given to using a unified stage-wise hardware characterization to guide the allocation of optimization effort between memory-bound and compute-bound stages. This motivates a coordinated end-to-end design.

\subsection{Roofline Performance Model}
The Roofline model~\cite{williams2009roofline, Koskelaroofline2018} is a principled analytical framework that bounds the attainable performance of an application on a given hardware architecture. By characterizing workloads through arithmetic intensity---the ratio of floating-point operations to memory traffic (FLOPs/Byte)---it establishes an upper performance bound and clearly distinguishes bandwidth-bound from compute-bound regimes.

In irregular workloads such as 3D Gaussian Splatting, pipeline stages exhibit highly heterogeneous arithmetic intensities. Front-end operations (e.g., attribute streaming and sorting) are typically bandwidth-bound, while back-end rasterization is compute-intensive. This stage-wise imbalance implies that end-to-end performance is highly sensitive to the optimization effort between reducing memory traffic and lowering arithmetic cost.

Although recent 3DGS accelerators~\cite{Gao2026SwiftGS, Axis-Shared_2026} employ the Roofline model to analyze and optimize individual kernels, few works leverage a quantitative hardware performance model to systematically determine \emph{which} stages should prioritize memory-traffic reduction versus arithmetic-cost reduction. RoofGS uses the Roofline model as a practical design guide for selecting and integrating stage-aware execution optimizations across the rendering pipeline.

\begin{figure*}[t]
    \centering
    \includegraphics[trim={2.0cm 1cm 0.75cm 0.5cm}, clip, width=1\linewidth]{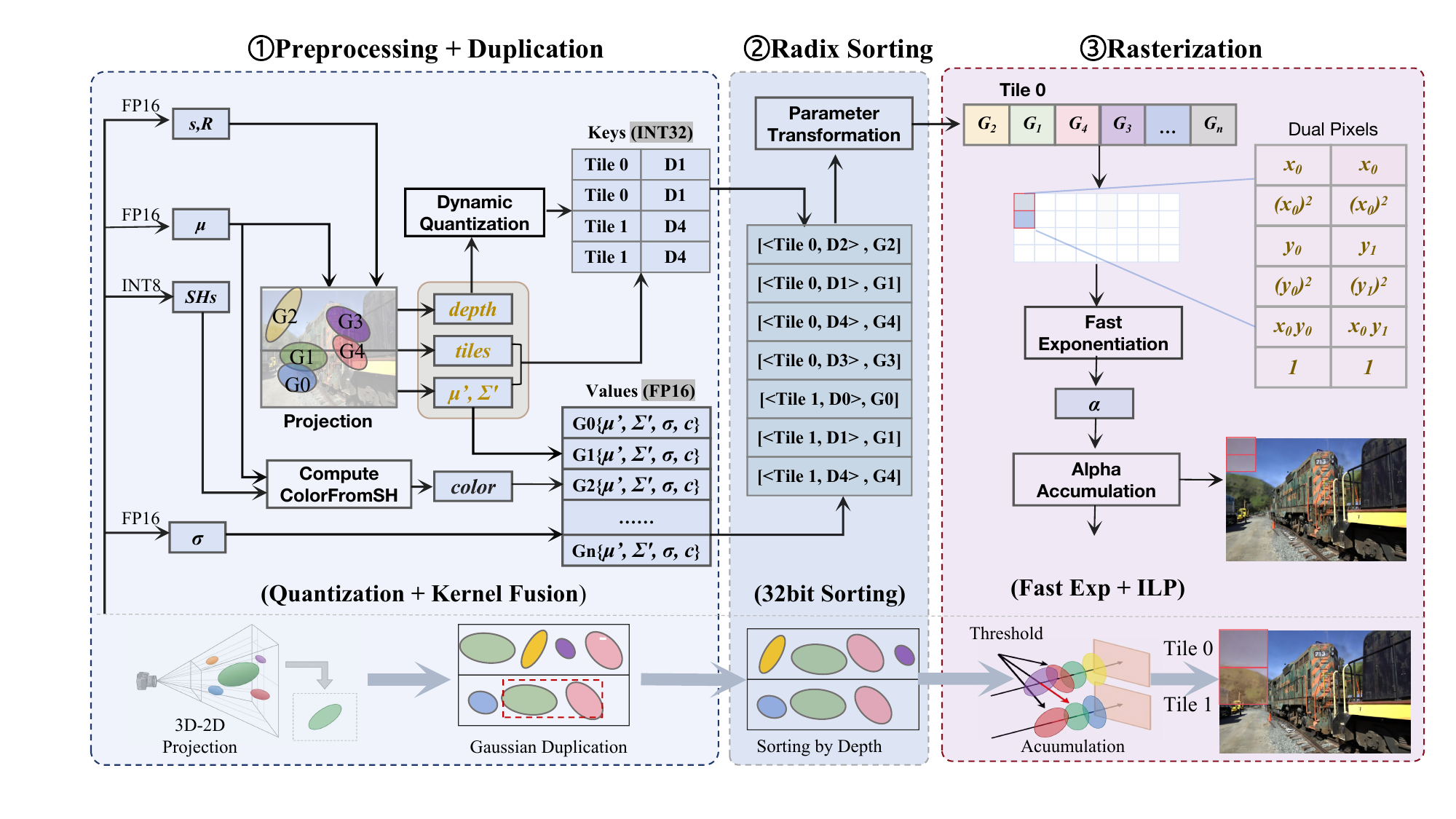}
    \captionsetup{
        font=small,
        labelfont=bf,
        justification=justified,
    }
\caption{Overall architecture of RoofGS. It streamlines the pipeline through quantization-aware kernel fusion in preprocessing, a compressed 32-bit radix sort, and hardware-accelerated rasterization using dual-pixel instruction-level parallelism.}
\label{fig:overview_framework}
\end{figure*}

\section{Analytical Roofline Characterization}
\label{sec:roofline_analysis}

\begin{table}[t]
\small
\centering
\caption{Notation used in Roofline characterization.}
\label{tab:notation}
\begin{tabular}{c p{5.5cm}}
\toprule
Symbol & Description \\
\midrule
$\boldsymbol{\mu}, \mathbf{r}, \mathbf{s}, \alpha$
& Gaussian mean, rotation, scale, and opacity. \\
\midrule
$\boldsymbol{\mu}', \Sigma'$
& Projected 2D mean and covariance matrix. \\
\midrule
$M, N_{\mathrm{dup}}$
& Average tile duplication factor and total duplicated instances. \\
\midrule
$S_{\mathrm{kv}}, N_{\mathrm{pass}}$
& Key-value entry size and number of radix-sort passes. \\
\midrule
$I=F/B$
& Arithmetic intensity, defined as operations divided by data movement. \\
\bottomrule
\end{tabular}
\end{table}

We use Roofline analysis~\cite{Koskelaroofline2018} to identify the dominant bottlenecks of different
stages in the 3DGS rendering pipeline. The achievable performance is bounded by
\begin{equation}
P = \min(P_{\mathrm{peak}}, BW \cdot I),
\end{equation}
where $P_{\mathrm{peak}}$ is the peak compute throughput, $BW$ is the memory
bandwidth, and $I$ is arithmetic intensity. On an NVIDIA RTX~4090 GPU, the
FP32--DRAM balance point is approximately~\cite{nvidia2022ada}
\begin{equation}
I_{\mathrm{knee}}
=
\frac{82.6\,\text{TFLOPS}}{1008\,\text{GB/s}}
\approx 81.9\ \mathrm{FLOPs/Byte}.
\label{eq:knee_point}
\end{equation}
Stages far below this point are generally more sensitive to memory traffic,
whereas stages closer to it benefit more from compute optimizations.

\subsection{Preprocessing and Attribute Evaluation}

We estimate the arithmetic intensity of preprocessing to interpret its measured
Roofline position. In the baseline FP32 configuration, preprocessing computes
the 3D covariance and view-dependent color online using degree-3 spherical
harmonics (SH). The geometry attributes (position, scale, rotation, and
opacity) occupy $44\,\text{Bytes}$ per Gaussian, while the SH coefficients
require $192\,\text{Bytes}$, giving a nominal input volume of
\begin{equation}
D_{\mathrm{pre}}^{\mathrm{in}}
=
44 + 16 \times 3 \times 4
=
236\,\text{Bytes/Gaussian}.
\label{eq:pre_input_volume}
\end{equation}

Point projection, covariance construction and transformation, SH evaluation,
and screen-space extent computation require an estimated
$400$--$800\,\mathrm{FLOPs}$ per Gaussian. The resulting input-streaming
arithmetic intensity is
\begin{equation}
I_{\mathrm{pre}}^{\mathrm{in}}
=
\frac{400\text{--}800}{236}
\approx
1.69\text{--}3.39\ \mathrm{FLOPs/Byte}.
\label{eq:pre_intensity}
\end{equation}
This estimation excludes output and intermediate traffic. The
estimated arithmetic intensity is far below the knee point of
$81.9\,\mathrm{FLOPs/Byte}$ defined in Eq.~\eqref{eq:knee_point}, suggesting
that preprocessing is memory-bound, consistent with the observations reported
in GCC and GScore~\cite{lee2024gscore, pei2025gcc}.

\subsection{Duplication and Key-Value Generation}

During tile assignment, each Gaussian is duplicated to all tiles intersecting
its screen-space footprint. Let $M$ denote the average number of intersected
tiles. Each duplicated instance stores a tile-depth key and a Gaussian index,
requiring $12\,\text{Bytes}$ in the baseline implementation. Thus, the output
traffic grows linearly with $M$:
\begin{equation}
B_{\mathrm{dup}} \geq 12M\ \text{Bytes}.
\label{eq:dup_intensity}
\end{equation}

Key generation requires minimal computation, but reading Gaussian attributes and writing key--value pairs incur substantial global memory traffic. Since the number of generated entries grows with the number of tiles overlapped by each Gaussian, larger footprints can further increase memory traffic.

\subsection{Radix-Based Depth Sorting}
Before rasterization, duplicated key--value entries are sorted by tile and
depth. GPU radix sort reads and writes the key--value array in multiple
passes~\cite{satish2009efficient}. Its memory traffic can be approximated as
\begin{equation}
B_{\mathrm{sort}}
\propto
S_{\mathrm{kv}} N_{\mathrm{dup}} N_{\mathrm{pass}},
\label{eq:sort_memory}
\end{equation}
where $S_{\mathrm{kv}}$ denotes the entry size, $N_{\mathrm{dup}}$ the number
of entries, and $N_{\mathrm{pass}}$ the number of sorting passes. Because each
pass performs limited computation but accesses the full array, GPU radix sort
is often memory-bound~\cite{stehle2017memory, lee2024gscore}. Smaller entries
and fewer passes can therefore reduce sorting traffic.

\subsection{Rendering}
Following the rasterization procedure of 3D Gaussian Splatting~\cite{Kerbl20233DGS}, each Gaussian primitive is first projected onto the image plane, yielding the projected mean $\boldsymbol{\mu}'$ and covariance $\boldsymbol{\Sigma}'$. For a pixel located at $\mathbf{x}$, the contribution of the Gaussian is quantified by
\begin{equation}
\begin{aligned}
\alpha(\mathbf{x})
&=
o
\exp\left(
-\frac{1}{2}
(\mathbf{x}-\boldsymbol{\mu}')^{T}
\boldsymbol{\Sigma}'^{-1}
(\mathbf{x}-\boldsymbol{\mu}')
\right),
\label{eq:alpha_computation}
\end{aligned}
\end{equation}
where $o$ denotes the opacity and the Gaussians are ordered from front-to-back according to their depth. Here, $\alpha(\mathbf{x})$ represents the effective opacity of the Gaussian at pixel $\mathbf{x}$. Evaluating the exponential function introduces substantial instruction overhead, as a single IEEE-754 compliant single-precision \texttt{exp} operation typically expands into 15 to 20 GPU assembly (SASS) instructions~\cite{nvidia_cuda_guide, wong2010demystifying}.
 Additionally, each Gaussian--pixel contribution is evaluated and composited sequentially within a thread. Prior roofline analyses~\cite{pei2025gcc, Axis-Shared_2026, Gao2026SwiftGS} have also shown that the rasterization stage of 3DGS is compute-bound.

\section{Methods}
\label{sec:methods}

Following the Roofline characterization, RoofGS organizes the rendering pipeline according to a stage-specific optimization principle: bandwidth-limited stages prioritize reductions in global memory traffic, whereas compute-intensive stages prioritize lower instruction cost and increased instruction-level parallelism (ILP)~\cite{wall1991limits, jouppi1989nonuniform}. As shown in Fig.~\ref{fig:overview_framework}, this principle is instantiated through front-end fusion, compact attribute storage, compressed radix-sort keys, and a reorganized rasterizer with dual-pixel ILP and fast exponential evaluation.

\subsection{Memory-Bound Optimization}

RoofGS reduces the dominant global memory traffic in bandwidth-bound stages through three optimizations (Fig.~\ref{fig:overview_framework}): low-bit Gaussian-attribute compression, dynamic depth-key compression, and cross-stage kernel fusion.

\subsubsection{Compact Attribute Storage}

Streaming full-precision Gaussian attributes incurs high memory bandwidth demand. For third-degree SHs, each Gaussian
stores $3(D_{\mathrm{SH}}+1)^2=48$ coefficients across the RGB channels,
requiring 192 bytes in single precision when $D_{\mathrm{SH}}=3$
\cite{Kerbl20233DGS}. As shown on the left side of Fig.~\ref{fig:overview_framework},
single-precision SH coefficients dominate the memory traffic of appearance
attributes. Quantizing all SH coefficients with a single shared
scale~\cite{Jacob_2018_CVPR}, however, is ineffective because their dynamic
ranges differ substantially across frequency bands. The larger low-frequency
coefficients dominate the shared scale, resulting in coarse quantization of
the smaller high-frequency coefficients; conversely, a scale tailored to the
high-frequency terms may clip the low-frequency terms. This range mismatch can
degrade both the base appearance and fine-grained view-dependent details.

To avoid the range mismatch of a single shared scale, RoofGS quantizes SH
coefficients using separate symmetric INT8 scales for each SH degree
$d \in \{0,1,2,3\}$ and color channel $c \in \{R,G,B\}$
~\cite{NEURIPS2022_adf7fa39}. This degree- and channel-specific scheme accommodates the distinct coefficient distributions while reducing SH storage and memory traffic. The quantized coefficients are stored in global memory and
dequantized on the fly during SH evaluation, as illustrated by the INT8 SH
path in Fig.~\ref{fig:overview_framework}:

\begin{equation}
\begin{aligned}
s^{\mathrm{SH}}_{d,c}
&=
\frac{\max_i |\mathrm{SH}_{i,d,c}|}{127}, \\
\mathrm{SH}^{\mathrm{int8}}_{i,d,c}
&=
\operatorname{round}
\left(
\frac{\mathrm{SH}_{i,d,c}}{s^{\mathrm{SH}}_{d,c}}
\right), \\
\widehat{\mathrm{SH}}_{i,d,c}
&=
s^{\mathrm{SH}}_{d,c}\mathrm{SH}^{\mathrm{int8}}_{i,d,c}.
\end{aligned}
\label{eq:quant}
\end{equation}

Here, $\mathrm{SH}_{i,d,c}$ denotes the original FP32 coefficient,
$\mathrm{SH}^{\mathrm{int8}}_{i,d,c}$ is its quantized value stored in global
memory, and $\widehat{\mathrm{SH}}_{i,d,c}$ is the reconstructed FP32
coefficient used for SH evaluation.

RoofGS further reduces preprocessing traffic by storing the rotation
$\mathbf{r}$, scale $\mathbf{s}$, and opacity $o$ in FP16, while retaining the
3D mean $\boldsymbol{\mu}$ in FP32 to preserve projection accuracy. The FP16
attributes are converted to FP32 in registers before geometric computation.
Together with INT8 SH coefficients, this reduces the per-Gaussian attribute
footprint from 236 to approximately 96 bytes, a 59.3\% reduction. 

All low-precision attributes are reconstructed to FP32 in registers before any computation, so subsequent SH evaluation, covariance projection, and quadric evaluation are performed in single precision.

\subsubsection{Resolution-Adaptive Sorting-Key Encoding}
\label{sec:sorting_key}

The sorting traffic $B_{\mathrm{sort}}$ in Eq.~\eqref{eq:sort_memory} scales
with both the key--value entry size $S_{\mathrm{kv}}$ and the number of radix
passes $N_{\mathrm{pass}}$. In the baseline implementation, each duplicated
Gaussian is sorted using a 64-bit composite key consisting of a 32-bit tile ID
and a 32-bit floating-point depth. This representation requires eight radix
passes and repeatedly streams the duplicated array, making depth sorting a
major memory-bandwidth bottleneck.

RoofGS compresses the sorting key into a 32-bit word that jointly encodes the
tile index and a quantized camera-space depth, as illustrated in
Fig.~\ref{fig:overview_framework}. This design reduces $S_{\mathrm{kv}}$ and
cuts $N_{\mathrm{pass}}$ from 8 to 4, thereby lowering the dominant sorting
traffic.

The 32-bit budget is dynamically partitioned between the tile index and depth
according to the current tile layout. Here, ``dynamic'' refers to this
resolution-dependent bit allocation, rather than frame-wise estimation of the
minimum and maximum depths of visible Gaussians. The quantization interval
$[z_{\mathrm{near}},z_{\mathrm{far}}]$ is predefined according to the known
scene extent and camera-frustum configuration. Gaussians outside this valid
interval are rejected during visibility and frustum culling before key
generation and therefore do not affect the quantization range or participate
in sorting.

For a valid projected Gaussian with camera-space depth $d$, the quantized
depth is computed as

\begin{equation}
q_d =
\operatorname{clamp}\left(
\left\lfloor
\frac{d-z_{\mathrm{near}}}
     {z_{\mathrm{far}}-z_{\mathrm{near}}}
\left(2^{b_d}-1\right)
\right\rfloor,
0,\,
2^{b_d}-1
\right),
\label{eq:depth_quant}
\end{equation}

where $b_d$ is the number of bits allocated to depth. The clamp operation is
used only as a safeguard against numerical round-off at the interval
boundaries, rather than to accommodate arbitrary out-of-frustum primitives.
The mapping is monotonically non-decreasing, so quantization cannot reverse
the order of Gaussians assigned to distinct depth bins. Ordering ambiguity is
limited to Gaussians mapped to the same bin, whose depth difference is smaller
than the quantization step
$\Delta_d=(z_{\mathrm{far}}-z_{\mathrm{near}})/(2^{b_d}-1)$.

Given a tile grid containing $N_t=T_xT_y$ tiles, the tile index requires
$b_t=\lceil\log_2N_t\rceil$ bits, leaving $b_d=32-b_t$ bits for depth. The tile
index occupies the most significant bits, while the quantized depth occupies
the least significant bits. Integer sorting therefore first groups duplicated
Gaussians by tile and then orders them by quantized depth within each tile.

For example, at $1920\times1080$ resolution with $16\times16$ tiles, the tile
grid contains $120\times68=8{,}160$ tiles. Since
$2^{12}<8{,}160\leq2^{13}$, the tile index requires 13 bits, leaving 19 bits
for depth and providing $2^{19}=524{,}288$ quantization levels. The allocation
automatically adapts to the rendering resolution: a larger tile grid uses more
bits for the tile index, whereas a smaller grid leaves more bits for depth.

This compression is consistent with the sorting-traffic model in
Eq.~\eqref{eq:sort_memory}: reducing the key--value representation lowers
$S_{\mathrm{kv}}$, while reducing the key width lowers
$N_{\mathrm{pass}}$. Since radix sorting repeatedly streams the key--value
array and is primarily memory-bandwidth-bound, the compressed 32-bit key
directly reduces the dominant memory traffic and also cuts the number of sorting passes.

\subsubsection{Preprocessing Kernel Fusion}

The baseline 3DGS pipeline executes projection, tile duplication, and sorting
key generation as separate kernels. This separation introduces global memory
round trips for intermediate data such as screen-space coordinates
$\boldsymbol{\mu}'_i$, depth $d_i$, conic parameters $\Sigma_i'^{-1}$, and
decoded color $c_i$. As shown in the fused front-end of
Fig.~\ref{fig:overview_framework}, RoofGS removes these unnecessary transfers by
combining preprocessing, tile duplication, and key-value generation into a single kernel.

Within the fused kernel, each thread computes the projected Gaussian attributes,
evaluates tile coverage, and directly emits the duplicated entries. Duplicate
allocation is handled cooperatively at the thread-block level using
shared-memory prefix sums and a single global atomic operation per block, which
keeps allocation mostly on-chip and avoids fine-grained atomic contention. Since
$\boldsymbol{\mu}'_i$, $d_i$, and $\Sigma_i'^{-1}$ are retained in registers or
shared memory, RoofGS eliminates intermediate global memory round trips between
front-end stages. The quantized depth is directly packed into the 32-bit sorting
key, and only the final color and key-value arrays are written back to global
memory, reducing front-end global memory transactions from five to two per
primitive.

To further reduce the duplication traffic in \eqref{eq:dup_intensity} and the
number of sorted entries in \eqref{eq:sort_memory}, RoofGS adopts the Gaussian
culling strategy from Speedy-Splat~\cite{speedy-splat}. This culling step removes
Gaussians with negligible contributions before duplication, avoiding unnecessary
generation of duplicated key-value entries. As a result, it reduces the effective
duplication factor $M$, the number of duplicated entries $N_{\mathrm{dup}}$, and
the global memory traffic in both duplication and sorting.

\subsection{Compute-Bound Optimization}
\subsubsection{Rasterization Reorganization for Dual-Pixel ILP}

As established in the Roofline characterization
(Sec.~\ref{sec:roofline_analysis}), the rasterization stage is primarily
bounded by the arithmetic throughput of the Streaming Multiprocessors (SMs).
RoofGS therefore optimizes this stage by reformulating the per-pixel Gaussian
evaluation and increasing instruction-level parallelism (ILP) inside the
blending loop.

For each Gaussian-pixel pair, the baseline rasterizer first evaluates the
Gaussian power as

\begin{equation}
\begin{aligned}
p(x,y) = -\frac{1}{2}\Big(
& a(x-u)^2 + b(x-u)(y-v) \\
& {}+ c(y-v)^2
\Big).
\end{aligned}
\label{eq:power}
\end{equation}

which corresponds to the exponent term in the standard Gaussian opacity
evaluation defined in Eq.~\eqref{eq:alpha_computation}. The unclamped opacity
contribution is then computed as
\begin{equation}
\tilde{\alpha}(x,y) = o \exp(p(x,y)),
\label{eq:opacity_unclamped}
\end{equation}
where \((x,y)\) is the target pixel coordinate, \((u,v)\) denotes the projected
2D Gaussian mean \(\boldsymbol{\mu}'\), \((a,b,c)\) are the elements of the
inverse 2D covariance matrix \(\Sigma'^{-1}\), and \(o\in(0,1]\) is the learned
Gaussian opacity. The final opacity used for alpha blending is
\begin{equation}
\alpha(x,y)=\min(0.99,\tilde{\alpha}(x,y)).
\label{eq:opacity_clamped}
\end{equation}
In the baseline implementation, the offsets \((x-u)\) and \((y-v)\) are
recomputed for every intersecting primitive, introducing repeated subtraction
and temporary register usage within the innermost rendering loop.

To reduce this instruction overhead, RoofGS expands the quadratic form and
separates pixel-dependent terms from Gaussian-dependent terms:
\begin{equation}
\begin{aligned}
& a(x - u)^2 + b(x - u)(y - v) + c(y - v)^2 \\
& \quad = ax^2 + bxy + cy^2 - (2au + bv)x \\
& \quad\quad - (2cv + bu)y + C_0,
\label{eq:expand}
\end{aligned}
\end{equation}
where
\begin{equation}
C_0 = au^2 + buv + cv^2 .
\label{eq:c0}
\end{equation}

During the shared-memory loading stage, RoofGS folds the constant factor
\(-0.5\) and the opacity term \(\log o\) into the Gaussian coefficients:
\begin{equation}
\begin{aligned}
    C_1 &= -0.5a,  & C_4 &= au + 0.5bv, \\
    C_2 &= -0.5b,  & C_5 &= cv + 0.5bu, \\
    C_3 &= -0.5c,  & C_6 &= -0.5C_0 + \log o .
\end{aligned}
\label{eq:coeff}
\end{equation}
With these folded coefficients, the folded exponent can be evaluated as
\begin{equation}
P(x,y) =
C_1 x^2 + C_2 xy + C_3 y^2 + C_4 x + C_5 y + C_6.
\label{eq:dot}
\end{equation}
Here,
\begin{equation}
P(x,y)=p(x,y)+\log o,
\label{eq:folded_exponent}
\end{equation}
and therefore the unclamped opacity contribution can be directly written as
\begin{equation}
\tilde{\alpha}(x,y)=\exp(P(x,y)).
\label{eq:folded_alpha}
\end{equation}

Since each active thread is assigned to a fixed pixel, the coordinate terms
\(x^2\), \(xy\), \(y^2\), \(x\), and \(y\) remain invariant across all Gaussians
processed by that thread. RoofGS hoists these terms outside the primitive loop
and keeps them in registers. As a result, the inner-loop exponent evaluation
becomes a regular FMA sequence using the pre-folded coefficients in
\eqref{eq:coeff}, removing repeated coordinate subtraction and improving
instruction scheduling.

RoofGS further exploits the spatial coherence between vertically adjacent
pixels by assigning each thread to process two pixels, \((x,y)\) and
\((x,y+1)\). Both pixels traverse the same Gaussian sequence within a tile and
therefore can reuse the same shared-memory attribute load. Let \(P(x,y)\) be the
folded exponent computed by Eq.~\eqref{eq:dot}. For the adjacent pixel
\((x,y+1)\), direct substitution gives
\begin{equation}
\begin{aligned}
P(x,y+1) ={}& C_1x^2 + C_2x(y+1) + C_3(y+1)^2 \\
& + C_4x + C_5(y+1) + C_6 .
\end{aligned}
\label{eq:dual_pixel_substitution}
\end{equation}
Therefore, the second exponent can be derived incrementally as
\begin{equation}
P(x,y+1) = P(x,y) + C_2x + C_3(2y+1) + C_5.
\label{eq:dual_pixel}
\end{equation}
This avoids re-evaluating the full dot product for the second pixel. The first
pixel uses the regular FMA path in \eqref{eq:dot}, while the second pixel is
computed through the lightweight update in \eqref{eq:dual_pixel}. Thus, the
same Gaussian coefficients are reused across two pixel evaluations with only a
small amount of additional register-level arithmetic.

This dual-pixel formulation also halves per-pixel shared-memory traffic, since one attribute load is reused for both vertically adjacent pixels.

By combining coefficient folding, register-resident coordinate terms, and dual-pixel incremental evaluation, RoofGS reduces redundant arithmetic
instructions and on-chip memory accesses in the innermost blending loop. This
improves instruction scheduling and ILP within the rasterization core, enabling
the compute-bound stage to operate closer to the peak arithmetic throughput of
the SMs.

\subsubsection{Range-Aware Fast Exponential Evaluation}

Gaussian blending requires evaluating an exponential function for every valid
Gaussian--pixel pair, which can become a computational bottleneck. In the
reorganized rasterizer, the learned opacity \(o\) is folded into the exponent:
\begin{equation}
P(x,y)=p(x,y)+\log o,
\label{eq:fast_exp_folded}
\end{equation}
so that the unclamped opacity contribution becomes
\begin{equation}
\tilde{\alpha}(x,y)=\exp(P(x,y)).
\label{eq:fast_exp_alpha}
\end{equation}
As derived in Appendix~\ref{app:exp_range}, opacity culling restricts the
exponent for all surviving Gaussian--pixel pairs to
\begin{equation}
P(x,y)
\in
[-\ln 255,\,0]
\approx
[-5.541,\,0].
\label{eq:fast_exp_range}
\end{equation}

This narrow, non-positive range is well suited to a fast bit-level
approximation of the exponential function.

Following Schraudolph's approximation~\cite{schraudolph1999fast}, whose
bit-level construction is derived in Appendix~\ref{app:bit_exp_derivation}, we
approximate the IEEE 754 single-precision representation of \(\exp(P)\) by

\begin{equation}
\begin{aligned}
I(P)
&=
\left\lfloor
P \cdot 2^{23} \cdot \log_2(e)
+
2^{23} \cdot (127-\sigma)
\right\rfloor, \\
\widehat{\exp}(P)
&=
\operatorname{reinterpret}_{\mathrm{float}}\!\bigl(I(P)\bigr).
\end{aligned}
\label{eq:fma_exp}
\end{equation}

where \(2^{23}\) is the mantissa scale, \(127\) is the
single-precision exponent bias, and \(\sigma=0.045677\) is a calibration
constant that reduces the approximation error. Here, \(I(P)\) denotes a
32-bit integer whose bit pattern is constructed to approximate the IEEE~754
single-precision encoding of \(\exp(P)\). The
\(\operatorname{reinterpret}_{\mathrm{float}}\) operation performs a bitwise
reinterpretation rather than a numerical integer-to-floating-point conversion.
Consequently, \(\widehat{\exp}(P)\) denotes the resulting approximation to
\(\exp(P)\).

For the calibration constant \(\sigma=0.045677\), the relative error of
\(\widehat{\exp}(P)\) over the bounded range \(P\in[-\ln 255,0]\) is at most
\(\varepsilon_{\max}\approx0.03\). Hence the approximated opacity satisfies
\(\hat{\alpha}=\alpha(1+\varepsilon)\) with \(|\varepsilon|\le\varepsilon_{\max}\).
By the error propagation analysis in Appendix~\ref{app:fast_exp_error}, this
multiplicative perturbation leads to a per-pixel color error bounded by
\(|\Delta C|_\infty\le\varepsilon_{\max}\approx0.03\) in continuous color
space, independent of the number of blended Gaussians.

The opacity used for alpha blending is then computed as
\begin{equation}
\alpha(x,y)
=
\min\left(0.99,\hat{\exp}(P(x,y))\right).
\label{eq:fast_exp_final_alpha}
\end{equation}
Thus, the transcendental operation \(\exp(P)\) is replaced by a small number of
arithmetic and bit-level operations. This reduces the computational overhead
of the Gaussian blending loop while maintaining sufficient accuracy over the
bounded exponent range.

\begin{table}[t]
\centering
\caption{Orthogonality verification on RTX 4090: FPS and speedup over 3DGS (1080p and 4K).}
\label{tab:ortho}
\footnotesize
\setlength{\tabcolsep}{2.0pt}
\begin{tabular}{l *{4}{c c}}
\toprule
& \multicolumn{2}{c}{3DGS~\cite{Kerbl20233DGS}} & \multicolumn{2}{c}{RoofGS} & \multicolumn{2}{c}{PUP 3D-GS~\cite{PUP_3D-GS2025}} & \multicolumn{2}{c}{PUP + RoofGS} \\
\cmidrule(lr){2-3} \cmidrule(lr){4-5} \cmidrule(lr){6-7} \cmidrule(lr){8-9}
Scene & 1080p & 4K & 1080p & 4K & 1080p & 4K & 1080p & 4K \\
\midrule
train      & 157 & 46 & 1478 & 530 & 642 & 160 & 3000 & 1217 \\
truck      & 167 & 52 & 1348 & 585 & 352 & 98  & 2673 & 997  \\
drjohnson  & 131 & 38 & 1294 & 586 & 576 & 158 & 3248 & 1371 \\
playroom   & 175 & 53 & 1570 & 699 & 633 & 164 & 3414 & 1408 \\
\midrule
Mean       & 158 & 47 & 1423 & 600 & 551 & 145 & 3084 & 1248 \\
Speedup    & 1.0$\times$ & 1.0$\times$ & 9.0$\times$ & 12.7$\times$ & 3.5$\times$ & 3.1$\times$ & 19.6$\times$ & 26.4$\times$ \\
\bottomrule
\end{tabular}
\end{table}

\begin{figure*}[t]
    \centering
    \includegraphics[trim={2.4cm 2.8cm 2.0cm 1.0cm}, clip, width=1\linewidth]{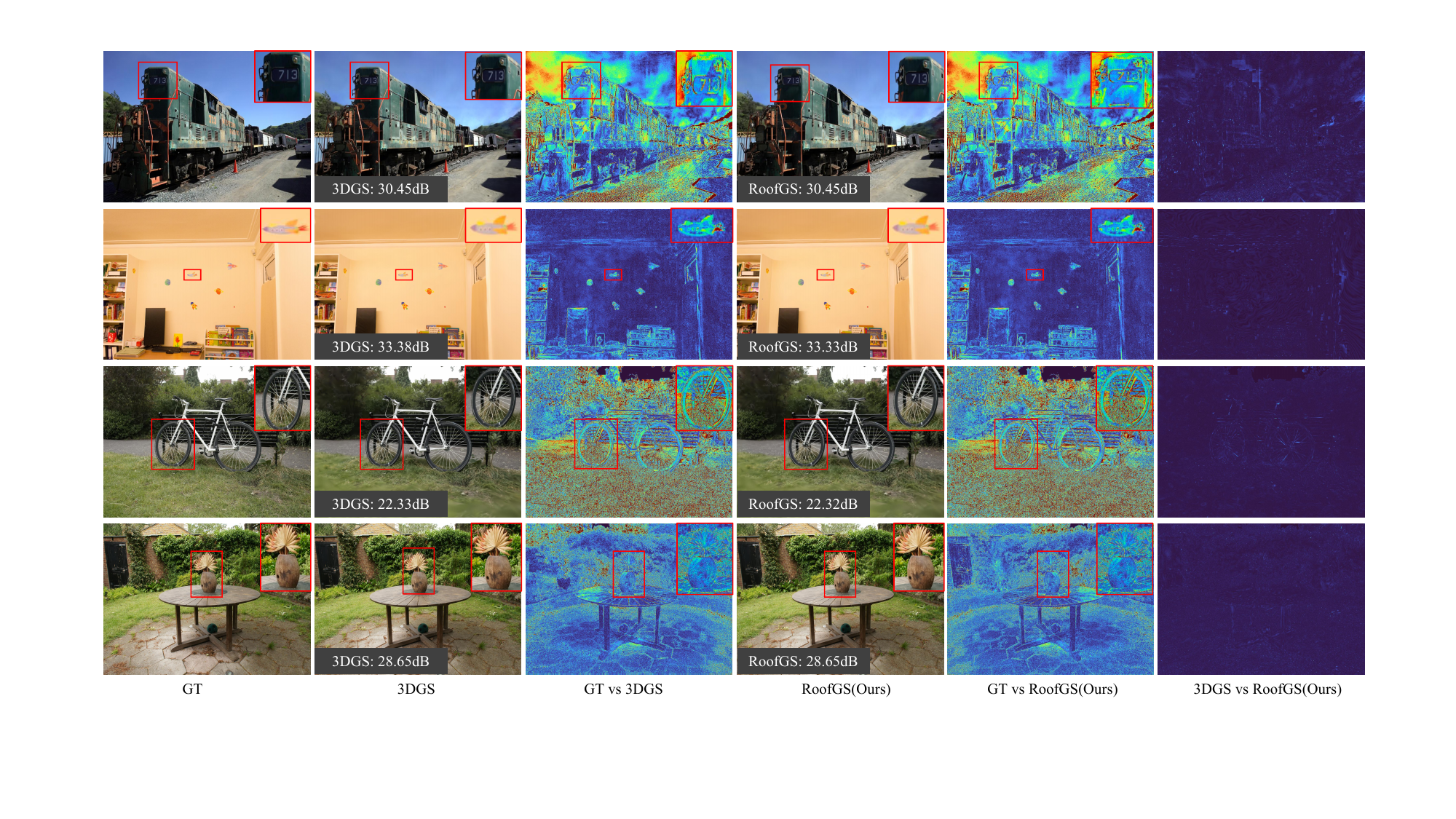}
    \captionsetup{
        font=small,
        labelfont=bf,
        justification=justified,
    }
\caption{\textbf{Qualitative comparison and residual error visualization.} 
Each row shows a representative test view. 
From left to right: Ground Truth (GT), baseline 3DGS, absolute residual map of 3DGS ($|\text{GT}-\text{3DGS}|$), RoofGS (Ours), absolute residual map of RoofGS ($|\text{GT}-\text{RoofGS}|$), and amplified difference map ($|\text{3DGS}-\text{RoofGS}|$). 
Red insets highlight zoomed-in regions containing fine details. RoofGS preserves reconstruction quality nearly identical to that of standard 3DGS. The amplified difference maps (rightmost column) confirm that residual deviations between the two methods are minimal and primarily confined to high-frequency edges.
Single-view quantitative metrics (PSNR) are overlaid on the rendered images. 
The right colorbar denotes the absolute RGB error scale $[0, 0.3]$.}
\label{fig:quality_compare}
\end{figure*}

\section{Conclusion}
\label{sec:conclusion}

We presented RoofGS, a Roofline-guided implementation framework for end-to-end 3D Gaussian Splatting rendering acceleration. Rather than applying a uniform optimization strategy across the pipeline, RoofGS uses stage-wise hardware characterization to distinguish bandwidth-sensitive front-end processing from computation-intensive rasterization. Based on this distinction, it combines front-end dataflow fusion, compact attribute and sorting-key representations, and rasterization reorganization with dual-pixel evaluation and fast exponential computation. Across the Mip-NeRF 360, Tanks \& Temples, and Deep Blending benchmarks at 4K resolution, RoofGS achieves an average $10.1\times$ speedup while incurring a PSNR reduction of only 0.028 dB, reaching 616 FPS on an RTX 4090. The results show that coordinating established optimization mechanisms according to stage-specific hardware constraints can provide substantial end-to-end gains for high-resolution 3DGS rendering. This perspective may also be applicable to future optimization of training pipelines and resource-constrained deployment settings.

\bibliography{ref}

@article{Kerbl20233DGS,
  author    = {Kerbl, Bernhard and Kopanas, Georgios and Leimkuehler, Thomas and Drettakis, George},
  title={{3D Gaussian Splatting} for Real-Time Radiance Field Rendering},
  journal   = {ACM Transactions on Graphics},
  volume    = {42},
  number    = {4},
  month     = {Jul.},
  year      = {2023},
  pages     = {85},
  doi       = {10.1145/3592433}
}

@inproceedings{mildenhall2020nerf,
  author    = {Mildenhall, Ben and Srinivasan, Pratul P. and Tancik, Matthew and Barron, Jonathan T. and Ramamoorthi, Ravi and Ng, Ren},
  title     = {{NeRF}: Representing Scenes as Neural Radiance Fields for View Synthesis},
  booktitle = {Proceedings of the European Conference on Computer Vision (ECCV)},
  series    = {Lecture Notes in Computer Science},
  volume    = {12346},
  pages     = {1--18},
  year      = {2020},
  publisher = {Springer},
  doi       = {10.1007/978-3-030-58452-8_1}
}

@ARTICLE{3dgsSurvey2025,
  author={Fei, Ben and Xu, Jingyi and Zhang, Rui and Zhou, Qingyuan and Yang, Weidong and He, Ying},
  journal={IEEE Transactions on Visualization and Computer Graphics}, 
  title={3D Gaussian Splatting as a New Era: A Survey}, 
  year={2025},
  volume={31},
  number={8},
  pages={4429--4449},
  doi={10.1109/TVCG.2024.3397828}}

@ARTICLE{RenderingReview2026,
  author={Yao, Mingyuan and Huo, Yukang and Ran, Yang and Tian, Qingbin and Wang, Ruifeng and Wang, Haihua},
  journal={IEEE Transactions on Visualization and Computer Graphics}, 
  title={Neural Radiance Field-Based Visual Rendering: A Comprehensive Review}, 
  year={2026},
  volume={32},
  number={7},
  pages={7361-7379},
  doi={10.1109/TVCG.2026.3677182}}

@ARTICLE{survey2024,
  author={Dalal, Anurag and Hagen, Daniel and Robbersmyr, Kjell G. and Knausgård, Kristian Muri},
  journal={IEEE Access}, 
  title={Gaussian Splatting: 3D Reconstruction and Novel View Synthesis: A Review}, 
  year={2024},
  volume={12},
  number={},
  pages={96797-96820},
  doi={10.1109/ACCESS.2024.3408318}}

@inproceedings{lee2024gscore,
  author    = {Lee, Junseo and Lee, Seokwon and Lee, Jungi and Park, Junyong and Sim, Jaewoong},
  title     = {{GSCore}: Efficient Radiance Field Rendering via Architectural Support for {3D} {Gaussian} {Splatting}},
  booktitle = {Proceedings of the 29th ACM International Conference on Architectural Support for Programming Languages and Operating Systems (ASPLOS)},
  volume    = {3},
  pages     = {497--511},
  year      = {2024},
  doi       = {10.1145/3620666.3651385}
}

@inproceedings{pei2025gcc,
  author    = {Pei, Minnan and Li, Gang and Si, Junwen and Zhu, Zeyu Susu and Mo, Zitao and Wang, Peisong and Song, Zhuoran and Liang, Xiaoyao and Cheng, Jian},
  title     = {{GCC}: A {3DGS} Inference Architecture with {Gaussian}-Wise and Cross-Stage Conditional Processing},
  booktitle = {Proceedings of the 58th IEEE/ACM International Symposium on Microarchitecture (MICRO)},
  pages     = {1824--1837},
  year      = {2025},
  doi       = {10.1145/3725843.3756072}
}

@inproceedings{zhu2026seele,
  author    = {Zhu, He and Huang, Xiaotong and Liu, Zihan and Lin, Weikai and Liu, Xiaohong and He, Zhezhi and Leng, Jingwen and Guo, Minyi and Feng, Yu},
  title     = {{Seele}: A Unified Acceleration Framework for Real-Time {Gaussian} {Splatting} on Mobile Devices},
  booktitle = {Proceedings of the IEEE/CVF Conference on Computer Vision and Pattern Recognition (CVPR)},
  pages     = {25979--25989},
  year      = {2026}
}

@inproceedings{liao2025tcgs,
  title={{TC-GS}: A Faster {Gaussian} {Splatting} Module Utilizing {Tensor} {Cores}},
  author={Liao, Zimu and Ding, Jifeng and Cui, Siwei and Gong, Ruixuan and Hu, Boni and Wang, Yi and Li, Hengjie and Wang, Hui and Zhang, Xingcheng and Fu, Rong},
  booktitle={Proceedings of the ACM SIGGRAPH Asia Conference Papers},
  pages={1--9},
  year={2025}
}

@article{li2026GEMMGS,
  title={GEMM-GS: Accelerating 3D Gaussian Splatting on Tensor Cores with GEMM-Compatible Blending},
  author={Li, Haomin and Zhu, Bowen and Liu, Fangxin and Wang, Zongwu and Liang, Xinran and Jiang, Li and Guan, Haibing},
  journal={arXiv preprint arXiv:2604.02120},
  year={2026}
}

@INPROCEEDINGS{speedy-splat,
  author={Hanson, Alex and Tu, Allen and Lin, Geng and Singla, Vasu and Zwicker, Matthias and Goldstein, Tom},
  booktitle={2025 IEEE/CVF Conference on Computer Vision and Pattern Recognition (CVPR)}, 
  title={Speedy-Splat: Fast 3D Gaussian Splatting with Sparse Pixels and Sparse Primitives}, 
  year={2025},
  volume={},
  number={},
  pages={21537-21546},
  doi={10.1109/CVPR52734.2025.02006}}

@inproceedings{wang2024adr,
  author    = {Wang, Xinzhe and Yi, Ran and Ma, Lizhuang},
  title     = {{AdR-Gaussian}: Accelerating {Gaussian} {Splatting} with Adaptive Radius},
  booktitle = {Proceedings of the SIGGRAPH Asia Conference Papers},
  year      = {2024},
  pages     = {1--10},
  publisher = {ACM},
  doi       = {10.1145/3680528.3687675}
}

@INPROCEEDINGS{FlashGS2025,
  author={Feng, Guofeng and Chen, Siyan and Fu, Rong and Liao, Zimu and Wang, Yi and Liu, Tao and Hu, Boni and Xu, Lining and Pei, Zhilin and Li, Hengjie and Li, Xiuhong and Sun, Ninghui and Zhang, Xingcheng and Dai, Bo},
  booktitle={2025 IEEE/CVF Conference on Computer Vision and Pattern Recognition (CVPR)}, 
  title={FlashGS: Efficient 3D Gaussian Splatting for Large-scale and High-resolution Rendering}, 
  year={2025},
  volume={},
  number={},
  pages={26652-26662},
  doi={10.1109/CVPR52734.2025.02482}}

@inproceedings{fan2024lightgaussian,
  title={LightGaussian: Unbounded 3D Gaussian Compression with 15x Reduction and 200+ {FPS}},
  author={Fan, Zhiwen and Wang, Kevin and Wen, Kairun and Zhu, Zehao and Xu, Dejia and Wang, Zhangyang},
  booktitle={Advances in Neural Information Processing Systems (NeurIPS)},
  year={2024}
}

@INPROCEEDINGS{Compressed3DGS2024,
  author={Niedermayr, Simon and Stumpfegger, Josef and Westermann, Rüdiger},
  booktitle={2024 IEEE/CVF Conference on Computer Vision and Pattern Recognition (CVPR)}, 
  title={Compressed 3D Gaussian Splatting for Accelerated Novel View Synthesis}, 
  year={2024},
  volume={},
  number={},
  pages={10349-10358},
  doi={10.1109/CVPR52733.2024.00985}}

@INPROCEEDINGS{PUP_3D-GS2025,
  author={Hanson, Alex and Tu, Allen and Singla, Vasu and Jayawardhana, Mayuka and Zwicker, Matthias and Goldstein, Tom},
  booktitle={2025 IEEE/CVF Conference on Computer Vision and Pattern Recognition (CVPR)}, 
  title={PUP 3D-GS: Principled Uncertainty Pruning for 3D Gaussian Splatting}, 
  year={2025},
  volume={},
  number={},
  pages={5949-5958},
  doi={10.1109/CVPR52734.2025.00558}}

@article{knapitsch2017tanks,
  title={{Tanks} and {Temples}: Benchmarking large-scale scene reconstruction},
  author={Knapitsch, Arno and Park, Jaesik and Zhou, Qian-Yi and Koltun, Vladlen},
  journal={IEEE Transactions on Visualization and Computer Graphics},
  volume={36},
  number={4},
  pages={1--13},
  year={2017},
  doi={10.1145/3072959.3073599}
}

@INPROCEEDINGS{Mip-NeRF-360,
  author={Barron, Jonathan T. and Mildenhall, Ben and Verbin, Dor and Srinivasan, Pratul P. and Hedman, Peter},
  booktitle={2022 IEEE/CVF Conference on Computer Vision and Pattern Recognition (CVPR)}, 
  title={Mip-NeRF 360: Unbounded Anti-Aliased Neural Radiance Fields}, 
  year={2022},
  volume={},
  number={},
  pages={5460-5469},
  doi={10.1109/CVPR52688.2022.00539}}

@article{williams2009roofline,
  author  = {Williams, Samuel and Waterman, Andrew and Patterson, David},
  title   = {{Roofline}: An Insightful Visual Performance Model for Multicore Architectures},
  journal = {Communications of the ACM},
  volume  = {52},
  number  = {4},
  pages   = {65--76},
  month   = apr,
  year    = {2009}
}

@InProceedings{ren2025fastgs,
  author    = {Ren, Shiwei and Wen, Tianci and Fang, Yongchun and Lu, Biao},
  title     = {FastGS: Training {3D} Gaussian Splatting in 100 Seconds},
  booktitle = {Proceedings of the IEEE/CVF Conference on Computer Vision and Pattern Recognition (CVPR)},
  month     = {June},
  year      = {2026},
  pages     = {26094--26103}
}

@ARTICLE{HAC++2025,
  author={Chen, Yihang and Wu, Qianyi and Lin, Weiyao and Harandi, Mehrtash and Cai, Jianfei},
  journal={IEEE Transactions on Pattern Analysis and Machine Intelligence}, 
  title={HAC++: Towards 100X Compression of 3D Gaussian Splatting}, 
  year={2025},
  volume={47},
  number={11},
  pages={10210-10226},
  doi={10.1109/TPAMI.2025.3594066}}

@inproceedings{Koskelaroofline2018,
  author    = {Koskela, Tuomas and Matveev, Zakhar and Yang, Charlene and Adedoyin, Adetokunbo and Belenov, Roman and Thierry, Philippe and Zhao, Zhengji and Gayatri, Rahulkumar and Shan, Hongzhang and Oliker, Leonid and Deslippe, Jack and Green, Ron and Williams, Samuel},
  editor    = {Yokota, Rio and Weiland, Mich{\`e}le and Keyes, David and Trinitis, Carsten},
  title     = {A Novel Multi-Level Integrated {Roofline} Model Approach for Performance Characterization},
  booktitle = {High Performance Computing},
  publisher = {Springer International Publishing},
  address   = {Cham, Switzerland},
  pages     = {226--245},
  year      = {2018}
}

@ARTICLE{compressionsurvey,
  author={Ali, Muhammad Salman and Zhang, Chaoning and Cagnazzo, Marco and Valenzise, Giuseppe and Tartaglione, Enzo and Bae, Sung-Ho},
  journal={IEEE Transactions on Circuits and Systems for Video Technology}, 
  title={Compression in 3D Gaussian Splatting: A Survey of Methods, Trends, and Future Directions}, 
  year={2026},
  pages={1-1},
  doi={10.1109/TCSVT.2026.3676048}}

@inproceedings{wang2026prune,
  author    = {Wang, Han and Huang, G. and Zhang, Fan and Bull, David and Anantrasirichai, Nantheera},
  title     = {Prune Wisely, Reconstruct Sharply: Compact {3D} {Gaussian} Splatting via Adaptive Pruning and Difference-of-{Gaussian} Primitives},
  booktitle = {Proceedings of the IEEE/CVF Conference on Computer Vision and Pattern Recognition (CVPR)},
  pages     = {11716--11725},
  year      = {2026}
}

@INPROCEEDINGS{Compact3d2024,
  author={Lee, Joo Chan and Rho, Daniel and Sun, Xiangyu and Ko, Jong Hwan and Park, Eunbyung},
  booktitle={2024 IEEE/CVF Conference on Computer Vision and Pattern Recognition (CVPR)}, 
  title={Compact 3D Gaussian Representation for Radiance Field}, 
  year={2024},
  volume={},
  number={},
  pages={21719-21728},
  doi={10.1109/CVPR52733.2024.02052}}

@INPROCEEDINGS{distill3dgs,
  author={Zhou, Shijie and Chang, Haoran and Jiang, Sicheng and Fan, Zhiwen and Zhu, Zehao and Xu, Dejia and Chari, Pradyumna and You, Suya and Wang, Zhangyang and Kadambi, Achuta},
  booktitle={2024 IEEE/CVF Conference on Computer Vision and Pattern Recognition (CVPR)}, 
  title={Feature 3DGS: Supercharging 3D Gaussian Splatting to Enable Distilled Feature Fields}, 
  year={2024},
  volume={},
  number={},
  pages={21676-21685},
  doi={10.1109/CVPR52733.2024.02048}}

@ARTICLE{Xuquantization2025,
  author={Xu, Hao and Wu, Xiaolin and Zhang, Xi},
  journal={IEEE Transactions on Image Processing}, 
  title={Improving 3D Gaussian Splatting Compression by Scene-Adaptive Lattice Vector Quantization}, 
  year={2026},
  volume={35},
  number={},
  pages={5864-5879},
  doi={10.1109/TIP.2026.3696119}}

@ARTICLE{Baosurvey2025,
  author={Bao, Yanqi and Ding, Tianyu and Huo, Jing and Liu, Yaoli and Li, Yuxin and Li, Wenbin and Gao, Yang and Luo, Jiebo},
  journal={IEEE Transactions on Circuits and Systems for Video Technology}, 
  title={3D Gaussian Splatting: Survey, Technologies, Challenges, and Opportunities}, 
  year={2025},
  volume={35},
  number={7},
  pages={6832-6852},
  doi={10.1109/TCSVT.2025.3538684}}

@ARTICLE{Visibility2026,
  author={Freitas, Davi R. and Tabus, Ioan and Guillemot, Christine},
  journal={IEEE Transactions on Multimedia}, 
  title={Visibility-Based Geometry Pruning of Neural Plenoptic Scene Representations}, 
  year={2026},
  volume={28},
  number={},
  pages={1-16},
  doi={10.1109/TMM.2025.3618548}}

@ARTICLE{POTR_2026,
  author={Ramlot, Bert and Courteaux, Martijn and Lambert, Peter and Wallendael, Glenn Van},
  journal={IEEE Transactions on Circuits and Systems for Video Technology}, 
  title={POTR: Post-Training 3DGS Compression}, 
  year={2026},
  pages={1-1},
  doi={10.1109/TCSVT.2026.3685779}}

@inproceedings{Labe2024DGD,
  author    = {Labe, Isaac and Issachar, Noam and Lang, Itai and Benaim, Sagie},
  title     = {{DGD}: Dynamic {3D} {Gaussians} Distillation},
  booktitle = {Computer Vision -- ECCV 2024},
  publisher = {Springer International Publishing},
  address   = {Cham, Switzerland},
  pages     = {361--378},
  year      = {2024},
  doi       = {10.1007/978-3-031-73113-6_21}
}

@article{hedman2018deep,
  title={Deep blending for free-viewpoint image-based rendering},
  author={Hedman, Peter and Philip, Julien and Price, True and Frahm, Jan-Michael and Drettakis, George and Brostow, Gabriel},
  journal={IEEE Transactions on Visualization and Computer Graphics},
  volume={37},
  number={6},
  pages={1--15},
  year={2018},
  doi={10.1145/3272127.3275084}
}

@article{schraudolph1999fast,
  author={N. N. Schraudolph},
  journal={Neural Computation},
  title={A Fast, Compact Approximation of the Exponential Function},
  year={1999},
  volume={11},
  number={4},
  pages={853--862},
  doi={10.1162/089976699300016467}
}

@INPROCEEDINGS{satish2009efficient,
  author    = {Satish, Nadathur and Harris, Mark and Garland, Michael},
  title     = {Designing Efficient Sorting Algorithms for Manycore GPUs},
  booktitle = {2009 IEEE International Symposium on Parallel \& Distributed Processing},
  year      = {2009},
  pages     = {1-10},
  doi       = {10.1109/IPDPS.2009.5161005}
}

@ARTICLE{robot_perception_2025,
  author={Zhou, Zhiyu and Hui, Feng and Wu, Yilin and Liu, Yu},
  journal={IEEE/ASME Transactions on Mechatronics}, 
  title={Six-DoF Pose Estimation With Efficient 3-D Gaussian Splatting Representation for Visual Relocalization}, 
  year={2025},
  volume={30},
  number={6},
  pages={4283-4292},
  doi={10.1109/TMECH.2024.3507134}}

@inproceedings{Gao2026SwiftGS,
  author    = {Lingjun Gao and Zhican Wang and Zhiwen Mo and Hongxiang Fan},
  title     = {{SwiftGS}: Algorithm and System Co-Optimization for Fast 3D Gaussian Splatting on GPUs},
  booktitle = {Proceedings of Machine Learning and Systems},
  editor    = {A. Chowdhery and Z. Jia},
  volume    = {8},
  pages     = {777--791},
  publisher = {Proceedings of Machine Learning and Systems},
  year      = {2026},
  url       = {https://proceedings.mlsys.org/paper_files/paper/2026/file/5f96a21345c138da929e99871fda138e-Paper-Conference.pdf}
}

@INPROCEEDINGS{Axis-Shared_2026,
  author={Wang, Zhican and He, Guanghui and Gao, Lingjun and Liu, Dantong and Hu, Shell Xu and Zhang, Chen and Song, Zhuoran and Lane, Nicholas and Fan, Hongxiang},
  booktitle={2026 ACM/IEEE 53rd Annual International Symposium on Computer Architecture (ISCA)}, 
  title={Efficient 3D Gaussian Splatting with Axis-Shared Rasterization and Order-independent Transmittance}, 
  year={2026},
  volume={},
  number={},
  pages={2014-2028},
  doi={10.1109/ISCA66397.2026.00144}}

@INPROCEEDINGS{stehle2017memory,
  author = {Stehle, Elias and Jacobsen, Hans-Arno},
  title = {{A Memory Bandwidth-Efficient Hybrid Radix Sort on GPUs}},
  booktitle = {Proceedings of the 2017 ACM International Conference on Management of Data (SIGMOD)},
  year = {2017},
  publisher = {ACM},
  address = {New York, NY, USA},
  doi = {10.1145/3035918.3035939}
}

@ARTICLE{Adversarial_2026,
  author={Shuai, Hui and Shi, Yucheng and Sun, Yubao and Liu, Qingshan},
  journal={IEEE Transactions on Multimedia}, 
  title={Adversarial Pruning Networks for Compact 3D Gaussian Splatting}, 
  year={2026},
  volume={28},
  number={},
  pages={1080-1089},
  doi={10.1109/TMM.2025.3632681}}

@ARTICLE{VINGS-Mono_2025,
  author={Wu, Ke and Zhang, Zicheng and Tie, Muer and Ai, Ziqing and Gan, Zhongxue and Ding, Wenchao},
  journal={IEEE Transactions on Robotics}, 
  title={VINGS-Mono: Visual-Inertial Gaussian Splatting Monocular SLAM in Large Scenes}, 
  year={2025},
  volume={41},
  number={},
  pages={5912-5931},
  doi={10.1109/TRO.2025.3613536}}

@ARTICLE{Duplex-GS_2026,
  author={Liu, Weihang and Li, Yuke and Li, Yuxuan and Yu, Jingyi and Lou, Xin},
  journal={IEEE Transactions on Circuits and Systems for Video Technology}, 
  title={Duplex-GS: Proxy-Guided Weighted Blending for Real-Time Order-Independent Gaussian Splatting}, 
  year={2026},
  volume={36},
  number={7},
  pages={9418-9431},
  doi={10.1109/TCSVT.2026.3666188}}

@InProceedings{Jacob_2018_CVPR,
author = {Jacob, Benoit and Kligys, Skirmantas and Chen, Bo and Zhu, Menglong and Tang, Matthew and Howard, Andrew and Adam, Hartwig and Kalenichenko, Dmitry},
title = {Quantization and Training of Neural Networks for Efficient Integer-Arithmetic-Only Inference},
booktitle = {Proceedings of the IEEE Conference on Computer Vision and Pattern Recognition (CVPR)},
month = {June},
year = {2018}
}

@inproceedings{wall1991limits,
  author    = {David W. Wall},
  title     = {Limits of Instruction-Level Parallelism},
  booktitle = {Proc. 4th Int. Conf. Architectural Support for Programming Languages and Operating Systems (ASPLOS IV)},
  pages     = {176--188},
  year      = {1991},
  doi       = {10.1145/106972.106991}
}

@article{jouppi1989nonuniform,
  author    = {Norman P. Jouppi},
  title     = {The Nonuniform Distribution of Instruction-Level and Machine Parallelism and Its Effect on Performance},
  journal   = {IEEE Transactions on Computers},
  volume    = {38},
  number    = {12},
  pages     = {1645--1658},
  year      = {1989},
  month     = dec,
  doi       = {10.1109/12.40844}
}

@inproceedings{NEURIPS2022_adf7fa39,
 author = {Yao, Zhewei and Yazdani Aminabadi, Reza and Zhang, Minjia and Wu, Xiaoxia and Li, Conglong and He, Yuxiong},
 booktitle = {Advances in Neural Information Processing Systems},
 doi = {10.52202/068431-1970},
 editor = {S. Koyejo and S. Mohamed and A. Agarwal and D. Belgrave and K. Cho and A. Oh},
 pages = {27168--27183},
 publisher = {Curran Associates, Inc.},
 title = {ZeroQuant: Efficient and Affordable Post-Training Quantization for Large-Scale Transformers},
 url = {https://proceedings.neurips.cc/paper_files/paper/2022/file/adf7fa39d65e2983d724ff7da57f00ac-Paper-Conference.pdf},
 volume = {35},
 year = {2022}
}

@misc{nvidia2022ada,
  title        = {NVIDIA {Ada} {GPU} {Architecture}},
  author       = {{NVIDIA Corporation}},
  year         = {2022},
  howpublished = {\url{https://images.nvidia.com/aem-dam/Solutions/geforce/ada/nvidia-ada-gpu-architecture.pdf}},
  note         = {Version V2.02. Appendix A contains GeForce RTX 4090 full specifications}
}

@INPROCEEDINGS{STREAMINGGS_2025,
  author={Zhang, Chenqi and Feng, Yu and Zhao, Jieru and Liu, Guangda and Ding, Wenchao and Wu, Chentao and Guo, Minyi},
  booktitle={2025 62nd ACM/IEEE Design Automation Conference (DAC)}, 
  title={STREAMINGGS: Voxel-Based Streaming 3D Gaussian Splatting with Memory Optimization and Architectural Support}, 
  year={2025},
  pages={1-7},
  doi={10.1109/DAC63849.2025.11132470}}

@inproceedings{du2026mobilegs,
  title     = {Mobile-GS: Real-time Gaussian Splatting for Mobile Devices},
  author    = {Du, Xiaobiao and Wang, Yida and Zhan, Kun and Yu, Xin},
  booktitle = {International Conference on Learning Representations (ICLR)},
  year      = {2026}
}

@inproceedings{xie2024mesongs,
  author    = {Xie, Shuzhao and Zhang, Weixiang and Tang, Chen and Bai, Yunpeng and Lu, Rongwei and Ge, Shijia and Wang, Zhi},
  title     = {{MesonGS}: Post-training Compression of {3D} Gaussians via Efficient Attribute Transformation},
  booktitle = {Computer Vision -- {ECCV} 2024},
  year      = {2024},
  publisher = {Springer Nature Switzerland},
  address   = {Cham},
  pages     = {421--438}
}

@InProceedings{Gao_2026_CVPR,
    author    = {Gao, Yuanyuan and Gong, Yuning and Liu, Yifei and Li, Jingfeng and Xu, Dan and Zhang, Yanci and Zhang, Dingwen and Sun, Xiao and Zhong, Zhihang},
    title     = {Proxy-GS: Unified Occlusion Priors for Training and Inference in Structured 3D Gaussian Splatting},
    booktitle = {Proceedings of the IEEE/CVF Conference on Computer Vision and Pattern Recognition (CVPR)},
    month     = {June},
    year      = {2026},
    pages     = {7330-7339}
}

@INPROCEEDINGS{OccluGaussian_2025,
  author={Liu, Shiyong and Tang, Xiao and Li, Zhihao and He, Yingfan and Ye, Chongjie and Liu, Jianzhuang and Huang, Binxiao and Zhou, Shunbo and Wu, Xiaofei},
  booktitle={2025 IEEE/CVF International Conference on Computer Vision (ICCV)}, 
  title={OccluGaussian: Occlusion-Aware Gaussian Splatting for Large Scene Reconstruction and Rendering}, 
  year={2025},
  volume={},
  number={},
  pages={26643-26652}
  }

@article{radl2024stopthepop,
  title={StopThePop: Sorted Gaussian Splatting for View-Consistent Rendering},
  author={Radl, Lukas and Kopanas, Michael and Kerbl, Bernhard and Steinberger, Markus and Schmalstieg, Dieter},
  journal={ACM Transactions on Graphics (TOG)},
  volume={43},
  number={4},
  pages={1--13},
  year={2024},
  publisher={ACM New York, NY, USA}
}

@manual{nvidia_cuda_guide,
  title        = {{NVIDIA CUDA C++ Programming Guide}},
  author       = {{NVIDIA Corporation}},
  year         = {2023},
  howpublished = {\url{https://docs.nvidia.com/cuda/cuda-c-programming-guide/}},
  note         = {Appendix G: Compute Capabilities \& Instruction Throughput. Accessed: 2024-03-15}
}

@inproceedings{wong2010demystifying,
  author    = {Wong, Henry and Papadopoulou, Maria-Margarita and Sadooghi-Alvandi, Maryam and Moshovos, Andreas},
  title     = {Demystifying GPU microarchitecture through microbenchmarking},
  booktitle = {2010 IEEE International Symposium on Performance Analysis of Systems \& Software (ISPASS)},
  pages     = {235--246},
  year      = {2010},
  publisher = {IEEE},
  doi       = {10.1109/ISPASS.2010.5452013}
}


\bibliographystyle{IEEEtran}

\end{document}